\documentclass[11pt,a4paper]{article}
\usepackage[hyperref]{emnlp2020}
\usepackage{times}
\usepackage{amsfonts}
\usepackage{amssymb}
\usepackage{graphics}
\usepackage{amsmath}
\usepackage{multirow}
\usepackage{enumitem}
\usepackage{graphicx} 
\graphicspath{figures/} 
\usepackage[ruled,vlined]{algorithm2e} 
\usepackage{textcomp}

\usepackage{latexsym}

\usepackage{microtype}

\aclfinalcopy 
\def\aclpaperid{1463}

\title{Recurrent Inference in Text Editing}

\author{Ning Shi \\
  Georgia Institute of Technology \\
  \texttt{ning.shi@gatech.edu} \\\And
  Ziheng Zeng \\
  University of Illinois\\
  \texttt{zzeng13@illinois.edu} \\\AND
  Haotian Zhang \\
  Learnable, Inc. \\
  \texttt{haotian.zhang@learnable.ai} \\\And
  Yichen Gong \\
  Horizon Robotics, Inc. \\
  \texttt{yichen01.gong@horizon.ai}}

\date{}

\makeatletter
\renewcommand\paragraph{%
  \@startsection{paragraph}{4}{\z@}{0.5mm}%
  {-0.3em}{\normalfont\normalsize\bfseries}}
\makeatother

\usepackage{fancyhdr}
\fancypagestyle{firststyle}
{
  \fancyhf{}
  \fancyfoot[C]{\footnotesize \textit{Findings of the 2020 Conference on Empirical Methods in Natural Language Processing}} 
  
}
\begin{document}
\maketitle
\begin{abstract}

In neural text editing, prevalent sequence-to-sequence based approaches directly map the unedited text either to the edited text or the editing operations, in which the performance is degraded by the limited source text encoding and long, varying decoding steps. To address this problem, we propose a new inference method, \textit{Recurrence}, that iteratively performs editing actions, significantly narrowing the problem space. In each iteration, encoding the partially edited text, Recurrence decodes the latent representation, generates an action of short, fixed-length, and applies the action to complete a single edit. For a comprehensive comparison, we introduce three types of text editing tasks: Arithmetic Operators Restoration (AOR), Arithmetic Equation Simplification (AES), Arithmetic Equation Correction (AEC). Extensive experiments on these tasks with varying difficulties demonstrate that Recurrence achieves improvements over conventional inference methods. 

\end{abstract}

\section{Introduction}

\begin{figure}[t!]
    \centering
    \includegraphics[width=\linewidth]{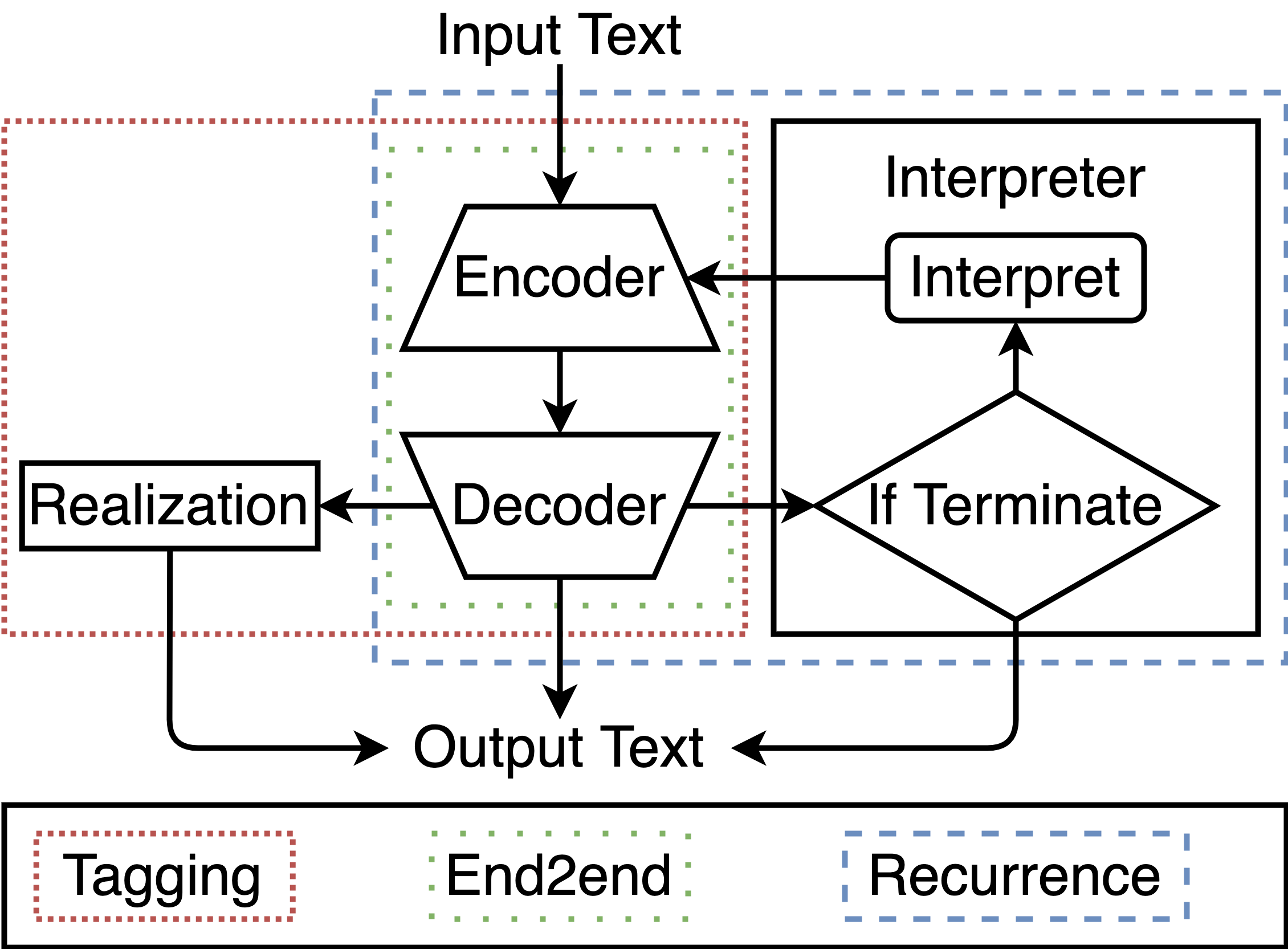}
    \caption{High-level illustration of End2end, Tagging, and Recurrence in text editing.}
    \label{fig: e2e_tag_rec_inference}
\end{figure}

For text editing, the sequence-to-sequence (seq2seq) framework has been applied to text simplification \cite{narayan-gardent-2014-hybrid, dong-etal-2019-editnts}, punctuation restoration \cite{tilk2016bidirectional, kim2019deep}, grammatical error correction \cite{ge-etal-2018-fluency, lichtarge2018weakly, zhao-etal-2019-improving}, machine translation post-editing \cite{libovicky-etal-2016-cuni,berard-etal-2017-lig}, and etc. We observe that current inference methods can be roughly grouped into two categories: \textit{End-to-end} (End2end) \cite{nisioi2017exploring, see-etal-2017-get, tan2017abstractive, junczys-dowmunt-etal-2018-approaching} and \textit{Tagging} \cite{filippova2015sentence, che2016punctuation,libovicky-etal-2016-cuni,wang-etal-2017-syntax, alva-manchego-etal-2017-learning,kim2019deep}. For models from both categories, the encoders extract and encode information from the source text sequence. Yet, the goal of the decoders is different for End2end and Tagging. Upon receiving the encoder's hidden states that comprise the source text information, the decoder of End2end directly decodes the hidden states and generates the completely edited target text sequence. But, the decoder of Tagging produces a sequence of editing operations, such as deletion and insertion, that is later applied to the source text to yield the edited text via a realization step \cite{malmi-etal-2019-encode}. The mechanisms of End2end and Tagging are illustrated in Figure \ref{fig: e2e_tag_rec_inference}. 

However, both End2end and Tagging are problematic because as decoding progresses, the divergence between the partially edited text and the original text grows, rendering the encoder hidden states less and less helpful for decoding the edited text or editing operations toward the end of the editing process; and as the number of decoding steps increases with edited text length, decoding the completely edited text or the full editing operation sequence becomes more and more demanding.

To tackle the aforementioned issues, we propose a recurrent inference method, \textit{Recurrence}, for text-editing with the encoder-decoder framework. 
Recurrence consists of two components as illustrated in Figure \ref{fig: e2e_tag_rec_inference}: (i) an encoder-decoder model, namely the \textit{programmer}; (ii) an \textit{interpreter}. 
For a given source sequence, the programmer determines an \textit{editing action} that consists of an editing operation with the tokens it needs and the position in the source sequence to apply the operation. After the interpreter executes the editing action, the partially edited text is again fed to the programmer to determine the next appropriate editing action. This process repeats until the programmer decides that no further editing is needed.  

Intuitively, Recurrence is advantageous because (i) as a novel recurrent inference process, it is not constrained by model structures and generally applicable; (ii) the programmer only produces one single editing step, easing the learning difficulty; (iii) the encoder hidden states are updated for each decoding step, providing faithful latent representations; (iv) the decoder outputs an editing action of fixed sequence length, alleviating the problem caused by long decoding steps. Empirically, through three text editing tasks, namely Arithmetic Operators Restoration (AOR), Arithmetic Equation Simplification (AES) and Arithmetic Equation Correction (AEC), we show that Recurrence is data-efficient and more resilient to the text sequence length and the vocabulary size.   

Our contributions are the followings: (1) we demonstrate that many text editing tasks can be solved by multiple inference steps recurrently; (2) we propose a novel recurrent inference method, Recurrence, 
for text editing that tears an editing task down into iterations of editing actions; (3) we design three easily reproducible, proof-of-concept text editing tasks, AOR, AES and AEC; (4) we exhibit that Recurrence outperforms End2end and Tagging in all three text editing tasks and is (i) less sensitive to longer sequences; (ii) less sensitive to larger vocab sizes; (iii) less data-hungry to achieve superior or competitive performances.

The code for three inference methods, text editing tasks, data generation, and experiments in this work is available at: \url{https://github.com/ShiningLab/Recurrent-Text-Editing}. 

\begin{figure*}[ht!]
    \centering
    \includegraphics[width=\linewidth]{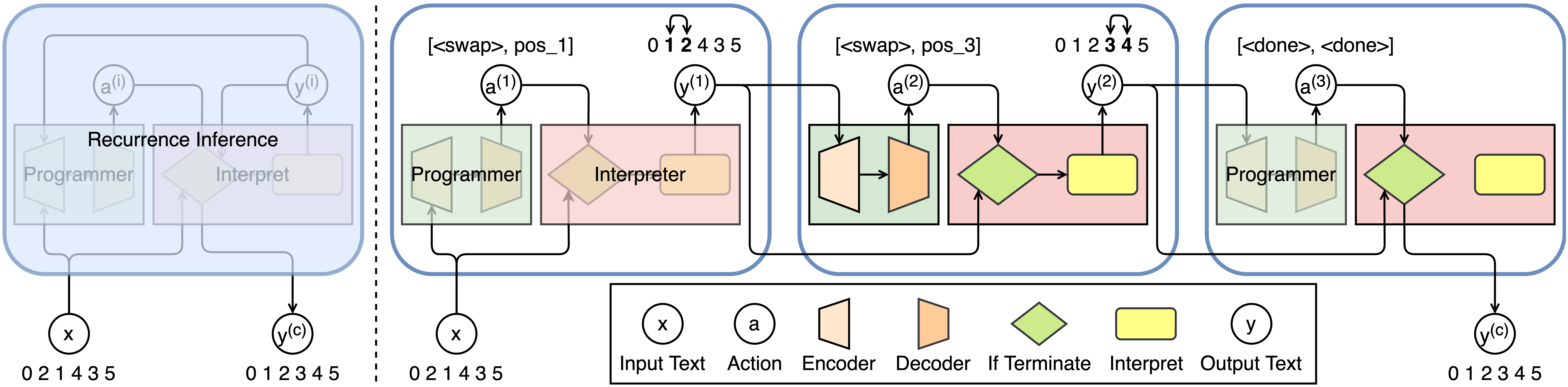}
    \caption{Illustrate Recurrence inference for text editing; the example shows an number ordering task where the number sequence $[0, 2, 1, 4, 3, 5]$ is edited to $[0, 1, 2, 3, 4, 5]$ via action $\mathbf{a}^{(1)}$, $[\textless\textnormal{swap}\textgreater, \textnormal{pos}\_1]$, which instructs the interpreter to swap number $1$ and $2$, and action $\mathbf{a}^{(2)}$, $[\textless \textnormal{swap}\textgreater, \textnormal{pos}\_3]$, which instructs the interpreter to swap number 3 and 4, imitating the bubble sort algorithm; finally, the interpreter halts inference and outputs the completely edited sequence $\mathbf{y}^{(\textnormal{c})}$ after receiving the termination action $\mathbf{a}^{(3)}=[\textless \textnormal{done}\textgreater, \textless \textnormal{done}\textgreater]$.}
    \label{fig: rec_inference}
\end{figure*}

\section{Related Work}

\paragraph{Text Editing} is an Natural Language Processing (NLP) task in that systems change texts by inserting, deleting and rephrasing the words to meet certain needs. According to the length relationship between input and output texts, we summarize text editing tasks into three types: \textit{short-to-long}, \textit{long-to-short}, and \textit{mixed}.

\paragraph{End-to-end} is one of the early methods to perform text editing by casting the job as seq2seq \cite{sutskever2014sequence} text generation. Without complicated preparation and subsequent processing, End2end has been proven to accomplish text editing well, in all three types \cite{tilk2016bidirectional,nisioi2017exploring,see-etal-2017-get,tan2017abstractive,junczys-dowmunt-etal-2018-approaching,zhao-etal-2019-improving}. Yet, conventional seq2seq-based approaches are well-known for their drawbacks, including dependency on large amounts of data, unexplainable processes, and uncontrollable outcomes \cite{wiseman-etal-2018-learning}. When texts do not need a complete modification, there are more appropriate methods than learning a direct mapping from unedited texts to edited texts.

\paragraph{Tagging} solves text editing in two steps instead. It firstly employs a seq2seq framework to produce tag sequences, and secondly, edits input texts according to the tag sequences (the ``realization" step) \cite{malmi-etal-2019-encode}. Tagging assigns the tag \texttt{KEEP} for words that do not need to be changed so that it does not need to learn a copy mechanism. Some have reported that Tagging is better than End2end in short-to-long \cite{che2016punctuation,kim2019deep}, long-to-short \cite{filippova2015sentence,alva-manchego-etal-2017-learning,wang-etal-2017-syntax}, and mixed editing \cite{libovicky-etal-2016-cuni,berard-etal-2017-lig,malmi-etal-2019-encode}. One notable member of the Tagging family is Neural Programmer-Interpreter (NPI), a recurrent and compositional neural network \cite{DBLP:journals/corr/ReedF15}. NPI is adopted in text editing to predict tags, such as \texttt{KEEP}, \texttt{DELETE}, and \texttt{INSERT}, and execute operations during decoding simultaneously. NPI-based methods have achieved state-of-the-art results in long-to-short \cite{dong-etal-2019-editnts,levenshtein-transformer}, and mixed editing \cite{vu-haffari-2018-automatic}. Nevertheless, like other Tagging methods, NPI's encoder hidden states are not updated during editing. Its decoder considers operations and executions from previous time steps to predict the current operation while putting massive pressure on the decoder \cite{hochreiter1998vanishing,bahdanau2014neural,cho2014properties}. Also, Tagging in general suffers from a performance decline caused by a large vocabulary that combines tags and words or too many decoding steps to assign tags. To resolve the aforementioned problems with Tagging, in Recurrence, we update the encoder hidden states iteratively and free the interpreter from the decoder to complete text editing in several program-interpret iterations (recurrent inference). NPI belongs to neural program induction \cite{10.5555/3305381.3305484}, but Recurrence is part of neural program synthesis \cite{NIPS2019_9116}. Consequently, Recurrence always follows the latest hidden representation of its input text rather than a static context matrix and only needs to decode an editing action of a fixed length in each iteration.

\paragraph{Multi-Step Learning} is a manner to solve a problem in several steps. Recent work in text editing prefers multi-step learning, especially for long-to-short \cite{narayan-gardent-2014-hybrid,zhang-lapata-2017-sentence}, and mixed editing \cite{ge-etal-2018-fluency,lichtarge2018weakly}. For example, Tagging can also be regarded as a two-steps learning. However, these studies usually edit texts incrementally through a multi-round seq2seq inference. To the best of our knowledge, our Recurrence is the first inference method that divides a text editing task into multiple independent sub-tasks and completes them recurrently.

\section{Recurrent Inference} \label{sec: method}

\subsection{Method Overview}

Recurrence breaks the text editing task down into iterations of editing actions and each editing action is determined on the hidden representation of the partially edited sequence. Conceptually, it is preforming a predefined underlying iterative algorithm that is designed to achieve some text editing goals. There are two components in Recurrence: programmer and interpreter. Given a source sequence $\mathbf{x} = x_1, \cdots, x_{|\mathbf{x}|}$, the programmer determines a single editing action, $\mathbf{a}^{(1)}$, to be applied on $\mathbf{x}$. Then, the interpreter executes the action $\mathbf{a}^{(1)}$ on $\mathbf{x}$ and produces the partially edited sequence with one edit, $\mathbf{y}^{(1)}$. Then, $\mathbf{y}^{(1)}$ is fed to the programmer to determine the next action $\mathbf{a}^{(2)}$. This process continues until the programmer determines the text is fully edited and outputs a termination action to stop further editing. The inference also ends if the number of iterations reaches a predefined limit. Finally, the interpreter outputs the completely edited sequence $\mathbf{y}^{(\textnormal{complete})}$. This recurrent editing process is illustrated with an example in the number ordering task in Figure \ref{fig: rec_inference}. 

The hypothesis is that it is easier to let a model learn a single editing step than the whole mapping between original and edited sequences. Also, being able to observe the latest text status leads to a more accurate input representation. Furthermore, Recurrence is explainable in the sense that not only we can understand the intention of each editing step done by the model, but we can also actively participate in designing the editing procedure.

\subsection{Programmer} 

Broadly speaking, the programmer determines the action for a given input in accordance with the underlying algorithm that the programmer is trained to mimic. In the programmer, the encoder extracts relevant information from an input text sequence $\mathbf{x}$ and then the decoder decides a single step of action that should be applied to $\mathbf{x}$. The programmer can be any model that is able to produce editing actions based on textual information. In our experiments, the programmer is a seq2seq model with an encoder-decoder architecture.

\begin{table*}[t!]
\centering
\resizebox{\textwidth}{!}{
\begin{tabular}{l}
\hline
\\
\hline
Input \\
Output \\ 
Input \\
Output \\ 
Input \\
Output \\
\hline
\end{tabular}

\begin{tabular}{l}
\hline
\textbf{AOR ($N=10$, $L=6$)}\\
\hline
$6\;\;10\;\;9\;\;5\;\;2\;\;3$ \\
$-\;6\;/\;10+9\;/\;5*2==3$ \\
$2\;\;2\;\;4\;\;8\;\;2\;\;4$ \\
$2*2-4+8\;/\;2==4$ \\
$2\;\;2\;\;4\;\;8\;\;2\;\;4$ \\
$-\;2+2\;/\;4*8+2==4$ \\
\hline
\end{tabular}

\begin{tabular}{l}
\hline
\textbf{AES} ($N=10$, $L=4$) \\
\hline
$-\;3+10\;/\;2==2$ \\
$-\;3+10\;/\;2==2$ \\
$(-\;2+4\;)\;/\;7*7==2$ \\
$2\;/\;7*7==2$ \\
$2\;/\;7*(\;11-4\;)==(\;4-2\;)$ \\
$2\;/\;7*7==2$ \\
\hline
\end{tabular}

\begin{tabular}{l}
\hline
\textbf{AEC} ($N=10$, $L=5$) \\
\hline
$4-3\;/\;6*4==2$ \\
$4-3\;/\;6*4==2$ \\
$6\;\;7*+\;/\;7+\;/\;7==2$ \\
$-\;7*5\;/\;7+7==2$ \\
$-\;6\;\;5+11-2$ \\
$-\;6+5+11-8==2$ \\
\hline
\end{tabular}
}
\caption{Examples from AOR, AES, and AEC with $N$ and $L$.} 
\label{tab: task_src_tgt_pairs}
\end{table*}

\subsection{Editing Actions}

An editing action contains (i) the type of editing operation, (ii) the position the editing occurs, and (iii) a text symbol.

Formally, the set of editing actions is defined by $\mathcal{A}:= \{\mathbf{a} = (e, p, s)|\forall e\in\mathcal{E},p\in\mathcal{P}, s\in\mathcal{S})\}$, where $\mathcal{E}$ is the set of all operations, $\mathcal{P}$ is the set of all positions, and $\mathcal{S}$ is the set of symbols. The definition of $\mathcal{E}$, $\mathcal{P}$ and $\mathcal{S}$ is determined by the specific text editing task and the underlying text editing algorithm. For example, each $p\in\mathcal{P}$ would contain a single position or multiple positions (i.e., a tuple of position indices) depending on the operation. Also, if an editing task contains only a single type of operation, then the operation can be omitted. Some operations, such as deletion, do not need a symbol input, so, the symbol component can also be omitted. It is required that $\texttt{DONE}\in\mathcal{E}, \mathcal{P},\mathcal{S}$ to indicate termination.

Editing actions allow the design of the editing order. Given $\mathbf{a}^{(1)},\cdots, \mathbf{a}^{(n)}$, the position sequence $p^{(1)}\in\mathbf{a}^{(1)},\cdots, p^{(n)}\in\mathbf{a}^{(n)}$ determines the editing order. This could be beneficial since empirical results have shown that ordering matters for text generation \cite{DBLP:conf/emnlp/FordD0D18}. For the sake of simplicity, in our experiments, we choose to arrange positions across actions in an increasing order, editing a sequence from left to right. 

Due to liberty given by the definition of the action, we believe Recurrence can be applied to a much border field of applications. In the scope of this paper, we only concern about text editing.

\subsection{Interpreter} 

The interpreter is a parameter-free function that executes the editing action produced by the programmer. Specifically, the interpreter first checks if the action is the termination action. If so, the interpreter will halt inference and directly output its input sequence as the completely edited text, $\mathbf{y}^{(\textnormal{complete})}$. Otherwise, the interpreter carries out the received action to its input sequence and produces a partially edited sequence. Then, the Recurrence continues by feeding the partially edited sequence into the programmer to determine the next editing action.

It is possible for the programmer to output illegal actions that do not follow the predefined action template (e.g., actions with missing a position component), especially when the programmer is not fully trained. Therefore, the interpreter checks if an action is valid and skips invalid actions by returning the input sequence.

\subsection{Offline Training} \label{subsec: offline_training}

Training text editing models requires pairs of source sequence $\mathbf{x}$ and target sequence $\mathbf{y}$, but different inference methods employ different generation algorithms to produce appropriate target sequences to form suitable training pairs. For the conventional inference methods, End2end map unedited text sequences to target text sequences directly, and Tagging map unedited text sequences to target tag sequences before realizing the target text sequences. Hence, for the training data, the source sequences are the original, while the target sequences are edited text sequences for End2end and editing operation sequences for Tagging. In our experiments, we name the training modes used by the conventional methods \textit{offline training}. 

\subsection{Online Training} \label{subsec: online_training}

To train the programmer, we compute all intermediate actions $\mathbf{a}^{(1)}, \cdots, \mathbf{a}^{(\textnormal{n})}$ that are required to edit input $\mathbf{x}$ to target $\mathbf{y^{(\textnormal{complete})}}$. Applying these editing actions, we obtain the partially edited sequences $\mathbf{y}^{(1)}=\mathbf{x}, \mathbf{y}^{(2)}, \cdots, \mathbf{y}^{(\textnormal{n})}=\mathbf{y}^{(\textnormal{complete})}$. After that, the training list of pairs for the programmer is $(\mathbf{y}^{(1)}, \mathbf{a}^{(1)}), (\mathbf{y}^{(2)}, \mathbf{a}^{(2)}), \cdots, (\mathbf{y}^{(\textnormal{n})}, \mathbf{a}^{(\textnormal{n})})$, where $\mathbf{a}^{(\textnormal{n})}$ is the termination action. We uniformly sample one source-target pair from this list as the training data instance. Due to the fact that selected training pairs for each source sequence $\mathbf{x}$ varies during training, we name this training mode \textit{online training}. For the thoroughness of experiments, we examine three inference methods with both training modes. In the training phase, intermediate training instances are exposed to $\textrm{End2end}_{\textnormal{Online}}$ and $\textrm{Tagging}_{\textnormal{Online}}$. Only the immediate editing action $(\mathbf{y}^{(1)}, \mathbf{a}^{(1)})$ are fed to $\textrm{Recurrence}_{\textnormal{Offline}}$.

\section{Tasks} \label{sec: task}

An increasing number of studies takes synthetic benchmark tasks to examine ideas before extending to open-domain natural language data \cite{DBLP:journals/corr/ZarembaS14,886a37b5fc2f43449e4bca3b5557e3ae,nangia-bowman-2018-listops,DBLP:conf/iclr/LampleC20}. Following the fruitful results of previous work, we aim to evaluate three inference methods in the domain of arithmetic problems \cite{hosseini-etal-2014-learning,roy-roth-2015-solving,ling-etal-2017-program} that can be treated as the test-beds for text editing. We introduce three tasks, namely AOR, AES, and AEC, corresponding to the three types of text editing tasks: short-to-long, long-to-short, and mixed. Being able to control the aspects of the datasets allows us to compare the characteristics of the three inference methods more thoroughly and analyze the appropriate situations to apply each method.

\subsection{Arithmetic Equation}

Our arithmetic equation consists of integer numbers $\mathcal{N}\in \mathbb{Z}^{\geq2}$, an equal sign (``$==$"), and operators\footnote{We use these symbols to apply the Python built-in function \texttt{eval()}.} $\mathcal{O} = \{``+",``-",``*",``/"\}$. For convenience, we restrict the right-hand side of the equation to a number. The equation holds if the value of the left-hand side equals the number on the right-hand side. Operators $\mathcal{O}$ are placed between two numbers, where the subtraction operator $``-"$ can also be put to the left of any single number. We consider equations as sequences of mathematical symbols \cite{DBLP:conf/iclr/SaxtonGHK19} instead of tree structures \cite{DBLP:conf/iclr/LampleC20}. We describe an arithmetic equation dataset from three aspects: 
(1) $N = |\mathcal{N}|$ defines the number of unique integers;
(2) $L\in\mathbb{Z}^*$ defines the number of integers in an equation;
(3) $D\in\mathbb{Z}^*$ defines the number of unique equations.

Note that since we only consider binary operations, the sequence length of a valid arithmetic expression is always $2L$ or $2L-1$, depending on if there is a subtraction operator before the first number. Intuitively, it is reasonable to assume that the greater $N$ and $L$ become, the harder the task gets. Whereas, the larger $D$, the easier the task becomes.

\begin{table*}[t!]
\centering
\resizebox{\textwidth}{!}{

\begin{tabular}{l}
\hline
 \\
\hline
Source \\
$\textrm{Target}_{\textrm{End2end}}$ \\
\shortstack[l]{$\textrm{Target}_{\textrm{Tagging}}$ \\ \vspace{0.25cm} \\ \vspace{0.25cm} } \\
$\textrm{Target}_{\textrm{Recurrence}}$ \\ 
\hline
\end{tabular}

\begin{tabular}{l}
\hline
\textbf{AOR} ($N=10$, $L=5$) \\
\hline
$8\;2\;8\;4\;2$ \\
$-\;8*2\;/\;8+4==2$ \\
\shortstack[l]{\textless insert\_$-$\textgreater \textless keep\textgreater \textless insert\_*\textgreater \textless keep\textgreater \\ \textless insert\_/\textgreater \textless keep\textgreater \textless insert\_+\textgreater \textless keep\textgreater \\ \textless insert\_==\textgreater  \textless keep\textgreater} \\
\textless pos\_0\textgreater \;$-$ \\
\hline
\end{tabular}

\begin{tabular}{l}
\hline
\textbf{AES} ($N=100$, $L=5$) \\
\hline
$-\;33+25+75-60==(\;30-23\;)$ \\
$-\;33+25+75-60==7$ \\ 
\shortstack[l]{\textless keep\textgreater \textless keep\textgreater \textless keep\textgreater \textless keep\textgreater \textless keep\textgreater \\ \textless keep\textgreater \textless keep\textgreater \textless keep\textgreater \textless keep\textgreater \textless sub\_7\textgreater \\ \textless delete\textgreater \textless delete\textgreater \textless delete\textgreater \textless delete\textgreater} \\
\textless pos\_9\textgreater \textless pos\_13\textgreater\;7 \\
\hline
\end{tabular}

\begin{tabular}{l}
\hline
\textbf{AEC} ($N=10$, $L=5$) \\
\hline
$7*8\;/\;4\;8\;2\;-==6$ \\
$7*8\;/\;4-8==6$ \\ 
\shortstack[l]{\textless keep\textgreater \textless keep\textgreater \textless keep\textgreater \textless keep\textgreater \\ \textless keep\textgreater \textless delete\textgreater \textless sub\_$-$\textgreater \textless sub\_8\textgreater \\ \textless keep\textgreater \textless keep\textgreater} \\
\textless delete\textgreater \textless pos\_5\textgreater \textless pos\_5\textgreater \\
\hline
\end{tabular}
}
\caption{Example target sequences given the same source sequence in AOR, AES, and AEC.} 
\label{tab: res_tgt}
\end{table*}

\subsection{Arithmetic Operators Restoration}  \label{sec: task-aor}

The goal of AOR is to convert a sequence of integer numbers into a valid arithmetic equation. For a given source sequence of integer numbers, $\mathbf{x}\in \mathcal{N}^{L}$, a model for AOR inserts appropriate operators from $\mathcal{O}$ in between the first $L-1$ integers in $\mathbf{x}$ and inserts an equal sign (``$==$") before the $L^{th}$ element in $\mathbf{x}$ so that the resulting arithmetic expression sequence (target sequence) is valid. Each integer sequence potentially corresponds to different valid arithmetic equations. Thus, AOR is \textit{one-to-many} learning. To obtain integer sequences for AOR, we first generate valid arithmetic equations and then remove all the operators and equal signs (see Table \ref{tab: task_src_tgt_pairs}). 

\subsection{Arithmetic Equation Simplification} \label{sec: task-aes}

Here, we involve two more mathematical symbols (``$($", ``$)$"). In an equation, parentheses help to group parts of an expression and indicate the order of precedence. In this task, we aim to simplify equations by calculating the parts in parentheses and removing parentheses from equations. Equation that has no parentheses is already in the simplest form, so there is no need to change. We generate complicated versions of a simplified equation by randomly replacing some integers (including the one on the right-hand side) with their equivalent bracketed expressions. Since these variants share the same simplified form, AES is \textit{many-to-one} learning (see Table \ref{tab: task_src_tgt_pairs}).

\subsection{Arithmetic Equation Correction} \label{sec: task-aec}

AEC is a more comprehensive text editing task in that a model needs to detect and correct possible mistakes. To generate mistakes, we inverse a valid equation by deleting, substituting, or inserting random tokens at random positions. We do not touch the right-hand side integer to guarantee that the corrected left-hand side (include ``==") equals the same value to assert equality. We fix the maximum number of errors to three, regardless the values of $N$, $L$, and $D$. No change is made if there is no error. We generate many wrong equations based on one correct equation. Meanwhile, a wrong equation can be modified into multiple correct equations. Hence, AEC is \textit{many-to-many} learning (see Table \ref{tab: task_src_tgt_pairs}). 

\section{Experiments} \label{sec: experiments}

We test Recurrence in comparison with End2end and Tagging across AOR, AES, and AEC. We describe the results conditioned on specific $N$, $L$, and $D$. Later, we analyze the impact of each of them in Section \ref{sec: analysis}.

\begin{table*}
\centering
\resizebox{\textwidth}{!}{
\begin{tabular}{ll}
\hline
 & \\
\hline
\textbf{Method} & \textbf{Training} \\
\hline
End2end      & Offline \\
 & Online \\
Tagging      & Offline \\
 & Online \\
Recurrence   & Offline \\
 & Online \\ 
\hline
\end{tabular}

\begin{tabular}{cc}
\hline
\multicolumn{2}{c}{\textbf{AOR} ($N=10$, $L=5$, $D=10$K)} \\
\hline
\textbf{\#Epoch} & \textbf{Equ Acc.\%} \\
\hline
3352 & 26.47 \\
2640 & 29.33 \\
1149 & 50.53 \\
2245 & 51.40 \\
1281 & 31.13 \\
1898 & $\textbf{58.53}^*$ \\ 
\hline
\end{tabular}

\begin{tabular}{ccc}
\hline
\multicolumn{3}{c}{\textbf{AES} ($N=100$, $L=5$, $D=10$K)} \\
\hline
\textbf{\#Epoch} & \textbf{Token Acc.\%} & \textbf{Seq Acc.\%} \\
\hline
5063 & 75.49 & 3.27 \\
7795 & 84.60 & 25.20 \\
5223 & 90.10 & 43.80 \\
4520 & 87.00 & 36.67 \\
7603 & 94.92 & 62.07 \\
7088 & $\textbf{98.63}^*$ & $\textbf{87.73}^*$ \\ 
\hline
\end{tabular}

\begin{tabular}{cccc}
\hline
\multicolumn{4}{c}{\textbf{AEC} ($N=10$, $L=5$, $D=10$K)} \\
\hline
\textbf{\#Epoch} & \textbf{Token Acc.\%} & \textbf{Seq Acc.\%} & \textbf{Equ Acc.\%}\\
\hline
72144 & 87.78 & 54.67 & 55.13 \\
112482 & $\textbf{88.08}^*$ & 57.27 & 57.73 \\
135729 & 82.29 & 44.20 & 44.40 \\
112968 & 84.46 & 46.93 & 47.33 \\
203067 & 81.85 & 55.87 & 56.20 \\
152982 & 83.64 & $\textbf{57.47}^*$ & $\textbf{58.27}^*$ \\ 
\hline
\end{tabular}
}
\caption{Evaluation results of three inference methods on AOR, AES, and AEC with specific $N$, $L$, and $D$.}
\label{tab: res_eva}
\end{table*}

\paragraph{Data.} 

In all tasks, the dataset is divided into three subsets: $70\%$ for training, $15\%$ for validation, and $15\%$ for testing. For AES (many-to-one learning) and AEC (many-to-many learning), we feed the training set to a data generator in every epoch to expose all the variants of targets as input sequences (see Section \ref{sec: task}). For the sake of fairness, we examine three methods in both online and offline training modes. To train $\textrm{End2end}_{\textrm{online}}$ and $\textrm{Tagging}_{\textrm{online}}$, in each epoch, we keep the targets, but uniformly pick a partially edited $\mathbf{y}^{(\textnormal{i})}$ to alternate the original input $\mathbf{x}$ as the source sequence. The target equations can be used to train End2end directly. By contrast, further pre-processing is necessary for Tagging and Recurrence. Training targets for Tagging are tag sequences, while those for Recurrence are editing actions.

\paragraph{Models.}

After testing Transformer \cite{NIPS2017_7181} and a range of modern RNNs \cite{mikolov2010recurrent,sutskever2014sequence,lecun2015deep}, we focus on the overall best-performed architecture --- bidirectional LSTM \cite{schuster1997bidirectional,hochreiter1997long} with an attention mechanism \cite{luong-etal-2015-effective}. Throughout all the experiments, three inference methods share the same model structure with $\textnormal{d}_{\textnormal{model}}=512$, $\textnormal{d}_{\textnormal{embedding}}=512$, $\textnormal{n}_{\textnormal{layers}}=1$, $\textnormal{r}_{\textnormal{learning}}=10^{-5}$, $\textnormal{r}_{\textnormal{teacher forcing}}=0.5$, and $\textnormal{r}_{\textnormal{dropout}}=0.5$ \cite{srivastava2014dropout}. Parameters are uniformly initialized from $[-\sqrt{\frac{1}{\textnormal{d}}}, \sqrt{\frac{1}{\textnormal{d}}}]$. To prevent uncontrolled interference, we train all models from scratch instead of pre-training. We use Adam optimizer \cite{DBLP:journals/corr/KingmaB14} with an L2 gradient clipping of $5.0$ \cite{10.5555/3042817.3043083}.

\paragraph{Evaluation.} 

We evaluate methods by three metrics: \textit{token accuracy}, \textit{sequence accuracy}, and \textit{equation accuracy}. Token accuracy marks the correct predictions at the token-level divided by the target sequence length and then averaged by the test size. Sequence accuracy stands for the correct predictions at the sequence-level divided by the test size. Equation accuracy is the number of true predicted equations divided by the test size; it emphasizes on whether an equation holds rather than whether an equation is the same as the target. We evaluate the performance via equation accuracy for AOR (one-to-many), sequence accuracy for AES (many-to-one), and both equation accuracy and sequence accuracy for AEC (many-to-many). Sequence accuracy is accompanied by token accuracy for additional reference.

\paragraph{Training.}

We train on a single GeForce RTX Titan with a batch size of 256. The last batch is dropped if it does not contain 256 samples. To ensure convergence, we adopt early stopping \cite{prechelt1998early} with a patience of 512 epochs.

\subsection{Arithmetic Operators Restoration} 

\paragraph{Data.}
Experiments are performed on a dataset with $N=10$, $L=5$, and $D=10$K. For Tagging, the tags are \texttt{KEEP} and $\texttt{INSERT}_{\texttt{TOKEN}_{\textnormal{AOR}}}$, where $\texttt{TOKEN}_{\textnormal{AOR}}=\mathcal{O}\bigcup\{``=="\}$. 

For Recurrence, the set of editing actions is defined as $\mathcal{A}_{\textnormal{AOR}} :=\{\mathbf{a} = (e, p, s)~|~\forall e\in\mathcal{E},p\in\mathcal{P}, s\in\mathcal{S})\}$, where $\mathcal{E}$ is an empty set since there is only one operation, insertion, and thus omitted; $\mathcal{P}:=\{p~|~p\in\{0,\cdots, |\mathbf{x}| \}\}$; and $\mathcal{S} = \texttt{TOKEN}_{\textnormal{AOR}}$. For a given action $\mathbf{a} = (p, s)$, the interpreter inserts $s$  before $x_p$ (see Table \ref{tab: res_tgt}).

\paragraph{Results.}

As shown in Table \ref{tab: res_eva}, $\textrm{Recurrence}_{\textrm{Online}}$ outperforms $\textrm{End2end}_{\textrm{Online}}$ by 29.20\% and $\textrm{Tagging}_{\textrm{Online}}$ by 7.13\%, achieving an equation accuracy of 58.53\%. Hence, $\textrm{Recurrence}_{\textrm{Online}}$ has the best performance. Note that online training is critical for $\textrm{Recurrence}$ to achieve good performance as  $\textrm{Recurrence}_{\textrm{Online}}$ outperforms $\textrm{Recurrence}_{\textrm{Offline}}$ by 27.40\%, whilst online training only helps to improve the performance of Tagging by 0.87\% and End2end by 2.86\%.  

\subsection{Arithmetic Equation Simplification}

\begin{figure}[htbp]
    \centering
    \includegraphics[width=\linewidth]{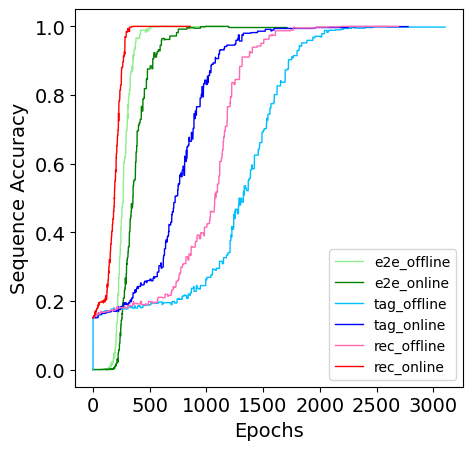}
    \caption{Testing sequence accuracy per epoch in AES with $N=10$, $L=5$, and $D=10$K; all methods achieved near-perfect performances, but $\textrm{Recurrence}_{\textnormal{Online}}$ converges the fastest.} 
    \label{fig: aes_10N}
\end{figure}

\paragraph{Data.}

We first experiment with $N=10$, $L=5$, and $D=10$K, but all methods can reach a near-perfect sequence accuracy (see Figure \ref{fig: aes_10N}). Therefore, we adjust $N$ from 10 to 100 to make the task more challenging. A target sequence to train Tagging is a sequence of tags consisting of \texttt{KEEP}, \texttt{DELETE}, and $\texttt{SUBSTITUTE}_{\texttt{TOKEN}_{\textrm{AES}}}$, where $\texttt{TOKEN}_{\textrm{AES}}\in\mathcal{N}$. For Recurrence, target editing actions are $\mathcal{A}_{\textnormal{AES}} :=\{\mathbf{a} = (e, p, s)~|~\forall e\in\mathcal{E},p\in\mathcal{P},s\in\mathcal{S})\}$, where the default operation is substitution, so $\mathcal{E}$ is an empty set and omitted; $\mathcal{P}:=\{\mathbf{p}=[p_1, p_2]~|~p_i\in\{0,\cdots, |\mathbf{x}| \}, \forall i=1,2\}$; $\mathcal{S}=\texttt{TOKEN}_{\textrm{AES}}$. This editing action instructs the interpreter to replace the part between $x_{p_1}$ and $x_{p_2}$ with $\texttt{TOKEN}_{\textrm{AES}}$ (see Table \ref{tab: res_tgt}).

\paragraph{Results.} 

Our $\textrm{Recurrence}_{\textrm{Online}}$ obtains a sequence accuracy of $87.73\%$, outperforming $\textrm{End2nd}_{\textrm{Online}}$ by $62.53\%$ and $\textrm{Tagging}_{\textrm{Offline}}$ by $43.93\%$. We also find that the performance of Recurrence is impaired significantly without online training. Besides, online training saves 515 epochs and achieves a better outcome. Both facts demonstrate the necessity of intermediate steps for training Recurrence.

\subsection{Arithmetic Equation Correction}

\paragraph{Data.} 
We use a dataset with $N=10$, $L=5$, and $D=10$K. A tag sequence is made of tags including \texttt{KEEP}, \texttt{DELETE}, $\texttt{SUBSTITUTE}_{\texttt{TOKEN}_{AEC}}$, and $\texttt{INSERT}_{\texttt{TOKEN}_{AEC}}$, where $\texttt{TOKEN}_{AEC}\in\mathcal{N}\bigcup\mathcal{O}\bigcup\{``=="\}$. For Recurrence, we define target editing actions as $\mathcal{A}_{\textrm{AEC}} :=\{\mathbf{a} = (e, p, s)~|~\forall e\in\mathcal{E},p\in\mathcal{P}, s\in\mathcal{S})\}$, where $\mathcal{E}:=\{e~|~e\in\{\texttt{DELETE},\texttt{SUBSTITUTE},\texttt{INSERT} \}\}$; $\mathcal{P}:=\{p~|~p\in\{0,\cdots, |\mathbf{x}| \}\}$; $\mathcal{S}:=\texttt{TOKEN}_{\textrm{AEC}}$. To freeze the sequence length of $\mathbf{a}$, we repeat $p$ at $\mathbf{a}_{3}$ to replace $s$ when $e=\texttt{DELETE}$.  During interpreting, $e=\texttt{DELETE}$ directs to remove $\mathbf{x}_{p}$; $e=\texttt{SUBSTITUTE}$ guides to replace $\mathbf{x}_{p}$ with $s$; $e=\texttt{INSERT}$ means to insert $s$ before $\mathbf{x}_{p}$ (see Table \ref{tab: res_tgt}).

\paragraph{Results.}
$\textrm{Recurrence}_{\textrm{Online}}$ attains higher scores over the other two methods, resulting in a sequence accuracy of 57.47\% and an equation accuracy of 58.27\%. The performance edge of Recurrence is not obvious due to the task setting. In section \ref{sec: analysis}, we adjust the task to distinguish the performance of each method more easily. When applying online training, we observe improvements in all three methods. Particularly, $\textrm{Recurrence}_{\textrm{Online}}$ takes around 50K epochs less than $\textrm{Recurrence}_{\textrm{Offline}}$ and attains a better performance.

\section{Analysis} \label{sec: analysis} 

\begin{figure*}[t!]
    \centering
    \includegraphics[width=\linewidth]{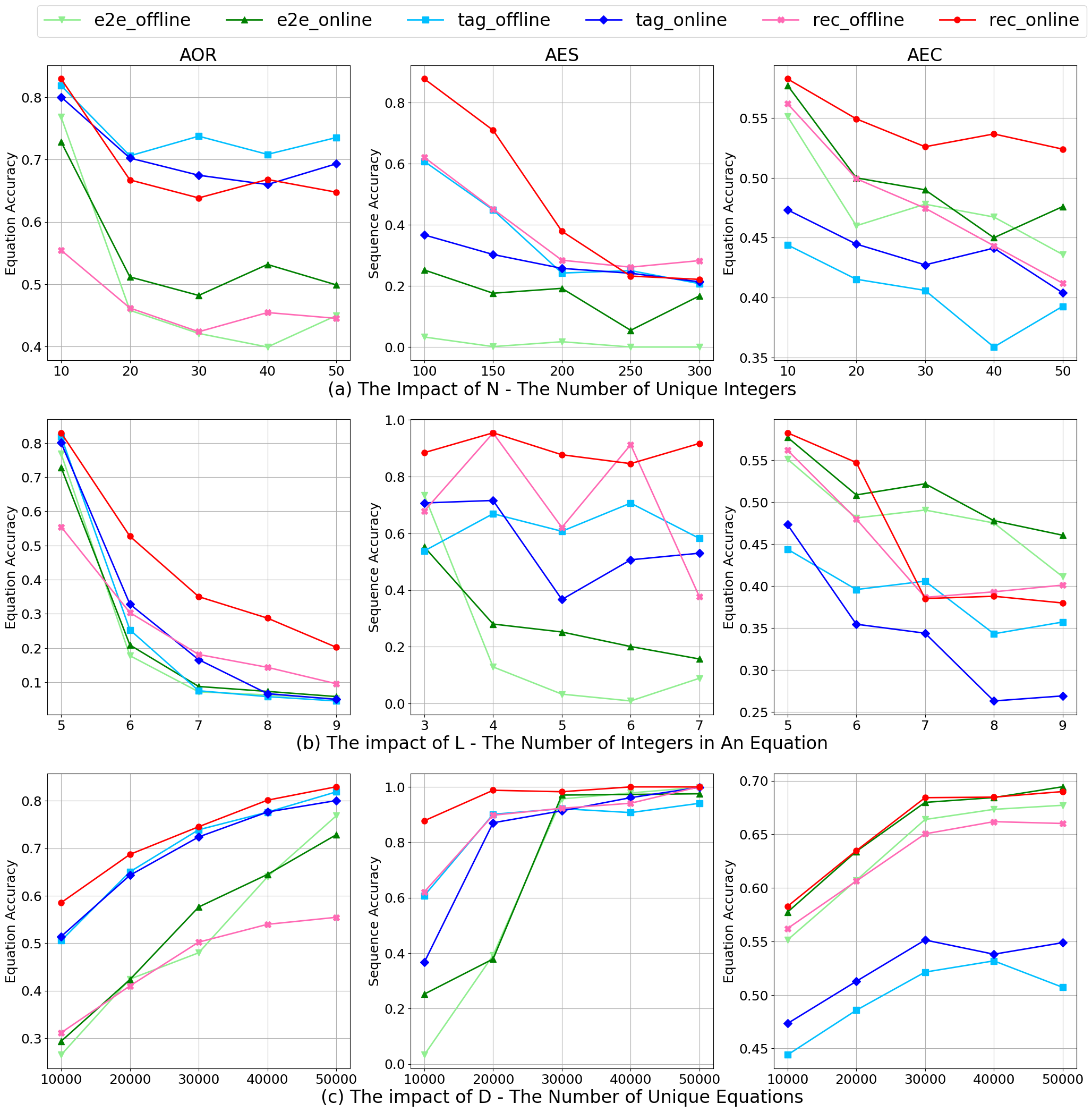}
    \caption{Evaluation results of three inference methods on AOR, AES, and AEC under the control of (a) $N$, (b) $L$, and (c) $D$, respectively.} 
    \label{fig: analysis}
\end{figure*}

As shown in section \ref{sec: experiments}, Recurrence outperforms End2end and Tagging in all three tasks in our experiment settings. In this section, we explore the limits of Recurrence by running experiments with varying values of $N$, $L$ and $D$, so as to determine in what scenario Recurrence performs well (see Figure \ref{fig: analysis}).

\paragraph{The Impact of $N$.} 

We conduct experiments with $L=5$, $D=50$K, and $N$ increasing from 10 to 50 with an interval of 10 for AOR;  $L=5$, $D=10$K, and $N$ increasing from 100 to 300 with an interval of 50 for AES; and  $L=5$, $D=10$K, and $N$ increasing from 10 to 50 with an interval of 10 for AEC. For AOR, $\textrm{Recurrence}_{\textnormal{Online}}$ and $\textrm{Tagging}$ show similar resilience, however, $\textrm{Tagging}_{\textnormal{Offline}}$ performs better when $N\geq 20$. For AES, $\textrm{Recurrence}_{\textnormal{Online}}$ performs much better than $\textrm{Tagging}$ and $\textrm{End2end}$ (by at least 20\%) when $N\leq 150$. Note that $\textrm{End2end}$ performs bad when $N\geq 100$ with $\textrm{End2end}_{\textnormal{Offline}}$ learns hardly anything. We also observe that $\textrm{End2end}_{\textnormal{Offline}}$ can achieve a near-perfect performance when $N=10$. These results indicate that the $\textrm{End2end}_{\textnormal{Offline}}$'s performance declines rapidly as $N$ increases and requires a much larger $D$-to-$N$ ratio to perform well. Finally, for AEC, $\textrm{Recurrence}_{\textnormal{Online}}$ displays the most resilience and performs the best.  

\paragraph{The Impact of $L$.}

We conduct experiments with $N=10$, $D=50$K, and $L$ increasing from 5 to 9 with an interval of 1 for AOR; $N=10$, $D=50$K, and $L$ increasing from 3 to 7 with an interval of 1 for AES; and settings identical to AOR for AEC. For AOR and AES, both $\textrm{Recurrence}_{\textnormal{Online}}$ and $\textrm{Tagging}$ show similar trend, however, $\textrm{Recurrence}_{\textnormal{Online}}$ performs the best. For AEC, while $\textrm{Recurrence}_{\textnormal{Online}}$ still outperforms $\textrm{Tagging}$, $\textrm{End2end}_{\textnormal{Offline}}$ performs the best for $L\geq 7$ and shows more resilience. We think when $N=10$, the AEC task is too easy for $\textrm{End2end}$ with $50$K training data. Thus, we increase $N$ from 10 to 100 and find that $\textrm{End2end}$ cannot gain any performance within 512 epochs (i.e., accuracy is 0\%). We want to stress that when the amount of data cannot counter the increase of $L$, which is the case for AOR and AES, $\textrm{End2end}$'s performance declines faster than $\textrm{Recurrence}$ and $\textrm{Tagging}$. 

\paragraph{The Impact of $D$.}

We conduct experiments with $N=10$, $L=5$, and $D$ increasing from 10K to 50K with an interval of 10K for AOR; $N=100$, $L=5$, and $D$ increasing from 10K to 50K with an interval of 10K for AES; and settings identical to AOR for AEC. All models benefit from the increasing of $D$ as expected. However, it is clear that $\textrm{Recurrence}_{\textnormal{Online}}$ is the best performing model when $D$ is small. The only exception is that for AEC, $\textrm{End2end}$ has similar performance trend as $\textrm{Recurrence}$. As discussed before, this is likely because \textrm{End2end} performs well with small $N$.

\paragraph{The Impact of Online Training.}

When comparing the performance between online and offline training, the online training, as expected, generally has better performances than offline training for End2end and Tagging with only a few exceptions. Note that online training is not part of the standard training procedure for End2end and Tagging, however, we use online training with End2end and Tagging for the sake of a fair comparison. Therefore, for End2end and Tagging, the online training acts like a data augmentation technique, providing more data points for training. Surprisingly, offline training also allows Recurrence to gain some editing ability, at times better than End2end and Tagging. We believe for text editing tasks with very localized editing actions, such as AES, showing the immediate editing actions are enough for the model to generalize proper editing actions. In other words, when the editing actions are less sequentially dependent, even offline training enables Recurrence to achieve performance better than End2end and Tagging. This supports our intuition that letting the programmer produce one single editing step reduces the learning difficulty.

\paragraph{The Impact of Ordering.} 

In early experiments, We find that the programmer cannot converge if the data guide it to edit a sequence in a random order (a mixture of both left-to-right and right-to-left). Hence, we think ordering matters for not only text generation \cite{DBLP:conf/emnlp/FordD0D18} but also Recurrence in text editing. One of our assumptions is that random ordering may assign various actions to the same text state, and thus causes confusion in the list of actions used to edit the input text $\mathbf{x}$ to the output text $\mathbf{y}$. When there are conflicting sample pairs in the training data set, the model cannot easily converge. We leave this problem for future work.

To summarize our findings, under settings with moderate or large $N$ and $L$, $\textrm{End2end}$ performs much worse than $\textrm{Tagging}$ and $\textrm{Recurrence}$ with limited data. $\textrm{Tagging}$ performs slightly better than $\textrm{Recurrence}$ when $N$ gets larger with fixed $D$ and $L$ in AOR (short-to-long). However, $\textrm{Tagging}$ performs worse than $\textrm{Recurrence}$ in all other cases. Therefore, we conclude that $\textrm{Recurrence}$ is more data-efficient and overall better performs than $\textrm{End2end}$ and $\textrm{Tagging}$ in most situations, especially in AES (long-to-short).

\section{Conclusions and Future Work}

We propose a recurrent inference method, Recurrence, that edits a given text sequence iteratively such that in each iteration the programmer determines a single step of editing action and the interpreter executes the action. Our method outperforms the other two inference methods, End2end and Tagging, in three arithmetic equation editing tasks we introduced. For future work, we plan to apply Recurrence to open-domain natural language data and investigate on how to relax its need for intermediate editing steps as extra supervision signals. We also wish to experiment with applying pointer attention \cite{NIPS2015_5866} to replace the position component in actions. 

\section{Acknowledgments} 
We give thanks to Keran Zhao, Yun He, Che Wang (Watcher) for sharing their pearls of wisdom. We also would like to thank EMNLP anonymous reviewers for their helpful insights, comments, and corrections. This research was supported by the Shining Lab.

\bibliography{emnlp2020}
\bibliographystyle{acl_natbib}

\end{document}


\appendix

\section{Recurrent Inference Algorithm}

We illustrate recurrent inference in Algorithm \ref{algo: recur_infer}, where $\mathbf{x}$ is the original text input; $\mathbf{a}^{(t)}$ is the action produced by the programmer in each iteration; $\mathbf{y}^{(t)}$ is the partially edited text and $\mathbf{y}^{(\textnormal{complete})}$ is the edited text. 

\begin{algorithm}
\SetAlgoLined
\KwResult{$\mathbf{y}^{(\textnormal{complete})}$}
 $\mathbf{x}_{\textnormal{Input}} = \mathbf{x}$\;
 $\textnormal{Terminate}=\textnormal{False}$\;
 $t =1$\;
 \While{Terminate is not True}{
  $\mathbf{a}^{(t)}=\texttt{Programmer}(\mathbf{x}_{\textnormal{Input}})$  \;
  $\mathbf{y}^{(t)}, \textnormal{Terminate}=\texttt{Interpreter}(\mathbf{x}_{\textnormal{Input}}, \mathbf{a}^{(t)})$ \;
  \If{Terminate is True}{
    $\mathbf{y}^{(\textnormal{complete})} = \mathbf{y}^{(t)}$\;
    break\;
  }
  $\mathbf{x}_{\textnormal{Input}} = \mathbf{y}^{(t)}$\;
  $t = t+1$\;
 }
 \caption{Recurrence}
 \label{algo: recur_infer}
\end{algorithm}

\newpage

\section{Training Details}

\paragraph{Training Time.} 

\begin{figure}[htbp!]
    \centering
    \includegraphics[scale=0.26]{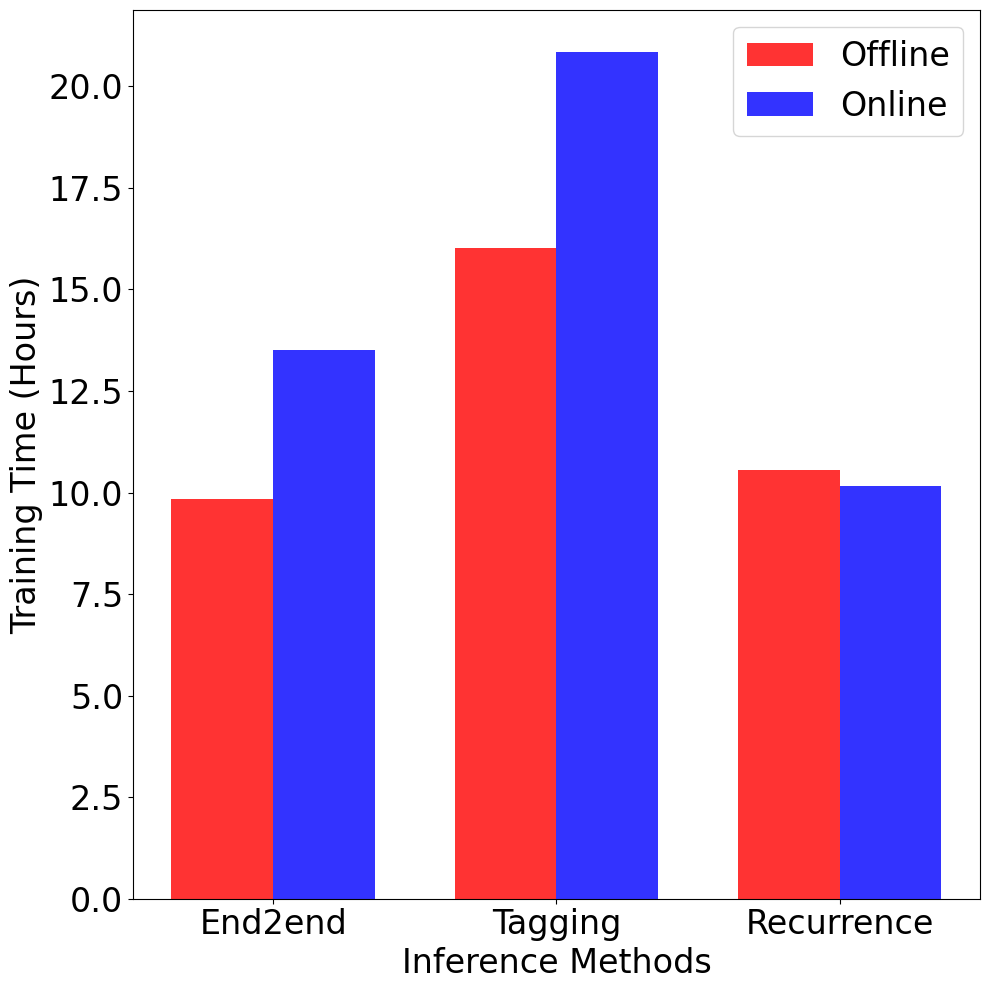}
    \caption{Average training time for End2end, Tagging, and Recurrence.}
    \label{fig: appendix_fig_2}
\end{figure}

Figure \ref{fig: appendix_fig_2} illustrates the average training time (in hours) over all experiments in this work for the three methods respectively. 

\paragraph{Training Progress.} 

For each training epoch, we keep track of performances on both validation set and test set (see Figure \ref{fig: appendix_fig_1}). 

 \noindent\begin{picture}(0,0)
\put(-235,-170){\begin{minipage}{\textwidth}
\centering
\includegraphics[scale=0.3]{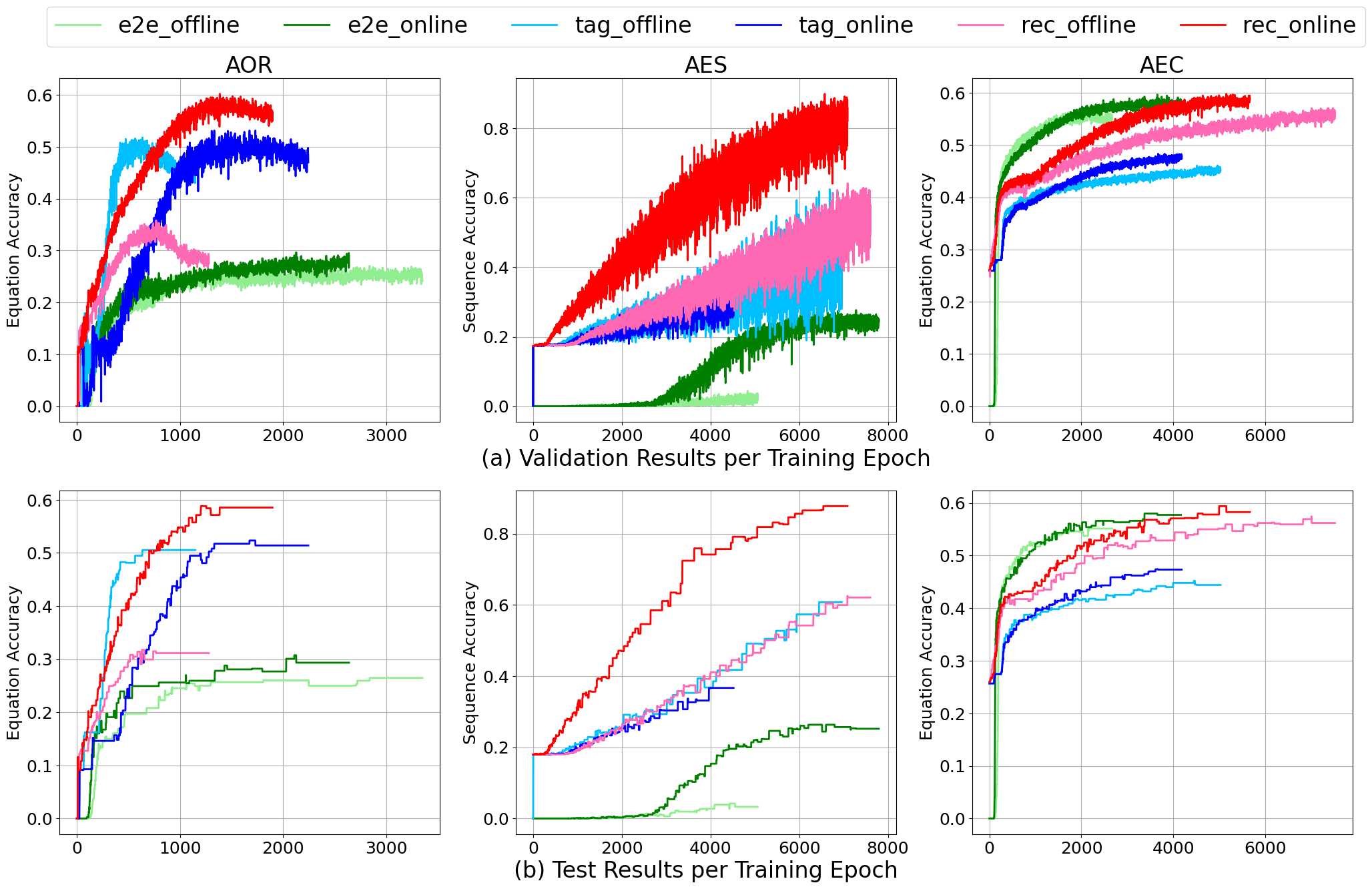}
\captionof{figure}{Validation and test results per training epoch for experiments in Section 5.}
\label{fig: appendix_fig_1}
\end{minipage}}
\end{picture}%